\documentclass{article}
\usepackage{spconf,amsmath,graphicx,hyperref}

\title{Activation Matching for Explanation Generation}
\name{Pirzada Suhail, Aditya Anand, Amit Sethi}
\address{IIT Bombay}
\begin{document}
%
\maketitle
\begin{abstract}

In this paper we introduce an activation-matching--based approach to generate minimal, faithful explanations for the decision-making of a pretrained classifier on any given image. Given an input image \(x\) and a frozen model \(f\), we train a lightweight autoencoder to output a binary mask \(m\) such that the explanation \(e = m \odot x\) preserves both the model's prediction and the intermediate activations of \(x\). Our objective combines: (i) multi-layer activation matching with KL divergence to align distributions and cross-entropy to retain the top-1 label for both the image and the explanation; (ii) mask priors---L1 area for minimality, a binarization penalty for crisp 0/1 masks, and total variation for compactness; and (iii) abductive constraints for faithfulness and necessity. Together, these objectives yield small, human-interpretable masks that retain classifier behavior while discarding irrelevant input regions, providing practical and faithful minimalist explanations for the decision making of the underlying model.

\end{abstract}
\section{Introduction}
\label{sec:intro}

Explanations are increasingly recognized as essential for understanding and trusting the decision-making of modern machine learning models. Deep neural networks, despite their remarkable predictive performance, often arrive at their outputs through complex, high-dimensional computations that are not directly human-interpretable. These models typically learn a vast repertoire of decision rules, any of which may be activated for a given input. As a result, simply observing the final prediction provides little insight into why the decision was made or which aspects of the input were most responsible.

Minimality has therefore emerged as a favored criterion for explanations. By isolating the smallest possible set of input features that suffices for a given prediction, one obtains an explanation that is both human-readable and faithful to the model’s internal computation. Minimal explanations highlight a compact subset of pixels in the case of images, or features in general, that directly support the output. Such explanations serve not only as cognitive aids for human understanding but also as a practical diagnostic tool: they can expose spurious correlations, highlight shortcut learning, and reveal when the model relies on inappropriate evidence. This is critical in safety-sensitive applications such as medical diagnostics, autonomous driving, and security, where knowing the precise basis for a decision can determine whether the system is trustworthy.

In this work, we propose an \emph{activation-matching} approach that, given an image and a frozen pretrained classifier, learns a lightweight autoencoder to produce a binary mask selecting a minimal set of pixels whose masked input preserves the model’s behavior providing interpretable understanding of the working of the machine learning model.

\section{Prior Work}

Inversion attempts to reconstruct inputs that elicit desired outputs or internal activations of a neural network. Unlike explanations, which are tied to a specific input and model decision, inversion focuses on synthesizing representative patterns that expose what a model has learned. Early studies on multilayer perceptrons applied gradient-based inversion to visualize decision rules, but these often yielded noisy or adversarial-like images~\cite{KINDERMANN1990277,784232,SAAD200778}. Evolutionary search and constrained optimization were explored as alternatives~\cite{Wong2017NeuralNI}. Later work introduced prior-based regularization, including smoothness constraints and pretrained generative models, to improve realism and interpretability of reconstructions~\cite{mahendran2015understanding,yosinski2015understanding,mordvintsev2015inceptionism,nguyen2016synthesizing,nguyen2017plug}. Recent advances include learning surrogate loss landscapes to stabilize inversion~\cite{liu2022landscapelearningneuralnetwork}, and generative methods that conditionally reconstruct inputs likely to produce a given output~\cite{suhail2024networkcnn}. Alternative formulations recast inversion into logical reasoning frameworks, encoding classifiers into CNF constraints for deterministic reconstruction~\cite{suhail2024network}.

While inversion aims to characterize model behavior in aggregate, explanation generation focuses on providing faithful rationales for a specific prediction. Explainable AI has therefore emerged as a major research area~\cite{ALI2023101805,hsieh2024comprehensiveguideexplainableai,Gilpin2018ExplainingEA}, motivated by the need to enhance trust, transparency, and accountability in high-stakes applications. Post-hoc attribution methods remain dominant: LIME produces local surrogate models~\cite{10022096}, Grad-CAM highlights salient image regions via gradient-weighted activations~\cite{Selvaraju_2019}, and more recent work emphasizes concept-based explanations that map predictions to semantically meaningful parts~\cite{10.1007/978-3-031-92648-8_17}. The quality of explanations is itself a key open challenge, with surveys stressing the need for rigorous metrics combining fidelity, stability, and human-centered evaluation~\cite{electronics10050593}. Explanations are also being integrated into interactive systems, allowing users to steer, debug, or refine models through explanation-guided feedback~\cite{teso2022leveragingexplanationsinteractivemachine}. Beyond heuristic methods, abductive reasoning approaches compute subset- or cardinality-minimal explanations with formal guarantees~\cite{ignatiev2018abductionbasedexplanationsmachinelearning}.

\section{Methodology}

We aim to generate minimal, faithful explanations for a frozen classifier \(f\) on any given image \(x\). 
To achieve this, we use a lightweight autoencoder that produces a binary mask \(m\). 
The masked explanation is defined as  

\[
e = m \odot x ,
\]

where irrelevant regions of the image are suppressed.  
The autoencoder is trained with a composite loss that combines activation-matching, fidelity, sparsity, binarization, smoothness, and robustness terms, each weighted appropriately to balance their contributions.  

\subsection{Activation matching and output fidelity}

The primary requirement is that the explanation \(e\) preserves the internal behavior of the classifier \(f\). 
Both the original image \(x\) and the explanation \(e\) are passed through the frozen model, and we enforce that their internal representations remain aligned.  

\subsubsection{Activation Matching Loss}
To ensure explanations activate the same computations as the original image, we minimize  

\[
\mathcal{L}_{\text{act}} = \sum_\ell \alpha_\ell \, d\!\big(\phi_\ell(x), \phi_\ell(e)\big),
\]

where \(\phi_\ell(\cdot)\) are post-ReLU activations, \(d\) is a distance function (MSE or cosine), and \(\alpha_\ell\) weights each layer.  

\subsubsection{KL Divergence Loss}
To align predictive distributions,  

\[
\mathcal{L}_{\text{KL}} = D_{\text{KL}}\!\big( \text{softmax}(f(x)) \,\|\, \text{softmax}(f(e)) \big).
\]

This encourages the explanation to reproduce the same class probabilities as the original image.  

\subsubsection{Cross-entropy Loss}
We further enforce that the explanation yields the same top-1 label as the original image using  

\[
\mathcal{L}_{\text{CE}} = - \log p_{f(e)}(y),
\]

where \(y\) is the predicted class from \(f(x)\).  

\subsection{Mask priors for minimality}

To ensure explanations are compact and human-interpretable, we impose priors on the mask \(m\).  

\subsubsection{Area Loss (sparsity)}
\[
\mathcal{L}_{\text{area}} = \|m\|_1 .
\]
This penalizes the active area of the mask, encouraging the use of as few pixels as possible.  

\subsubsection{Binarization loss (crispness)}
\[
\mathcal{L}_{\text{bin}} = \|m - m^2\|_1 .
\]
This drives mask values toward 0 or 1, ensuring sharp explanations rather than soft heatmaps.  
To enable backpropagation through the non-differentiable binarization step, we adopt a \emph{straight-through estimator} (STE), which treats the binary threshold operation as identity during the backward pass.  

\subsubsection{Total variation loss (smoothness)}
\[
\mathcal{L}_{\text{tv}} = \sum_{i,j} \big(|m_{i,j} - m_{i+1,j}| + |m_{i,j} - m_{i,j+1}|\big).
\]
This suppresses noisy, isolated activations, encouraging smooth and contiguous regions.  

\subsection{Abductive constraint}

Minimality alone does not guarantee sufficiency.  
We therefore add a robustness constraint: random perturbations outside the explanation should not change the prediction.  

Given a perturbed background \(r\), we construct  

\[
\tilde e = m \odot x + (1-m) \odot r ,
\]

and enforce  

\[
\mathcal{L}_{\text{rob}} = - \log p_{f(\tilde e)}(y),
\]

where \(y\) is the predicted class from the original image.  
This ensures the explanation remains valid even when irrelevant pixels are randomized.  

Together, these objectives ensure that the generated explanations are minimal, crisp, and robust, while faithfully capturing the classifier’s decision process.

\begin{figure*}[!t]
    \centering
    \includegraphics[width=0.925\linewidth]{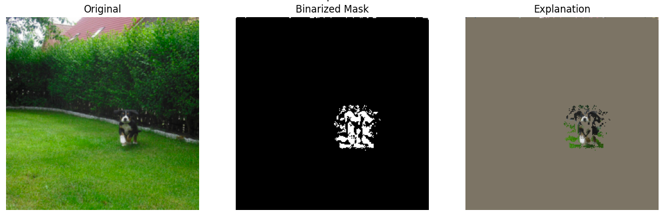}
    \caption{Original Image, 0/1 Mask, and Explanation}
    \label{fig:results}
\end{figure*}

\begin{figure}
    \centering
    \includegraphics[width=0.75\linewidth]{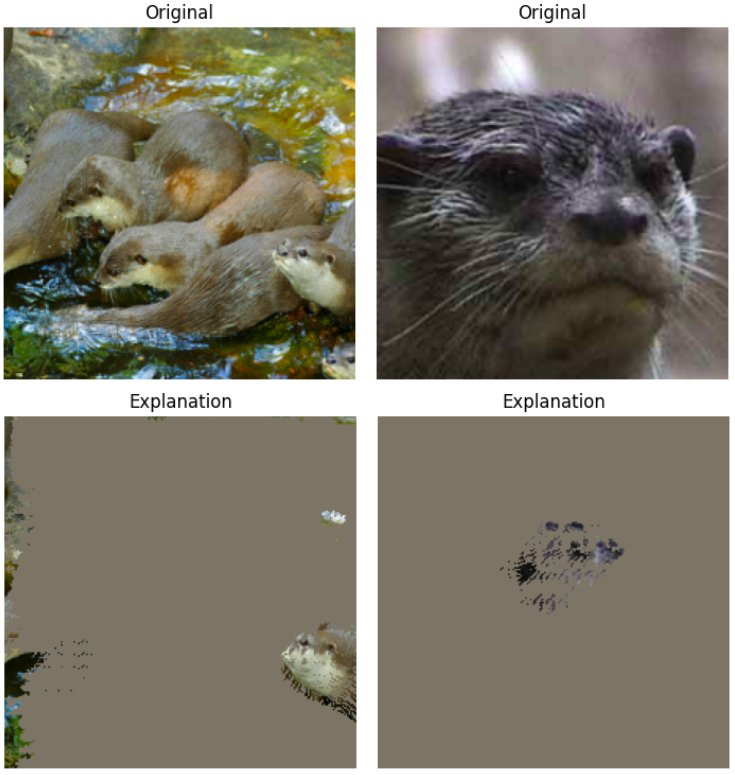}
    \caption{Explanations for sample Images of \emph{Otter}. 
    }
    \label{fig:results_otters}
\end{figure}

\section{Results}
\label{sec:pagestyle}

While our approach is general, we use it to explain the decision-making of a pretrained ResNet-18 classifier on ImageNet images. We define a simple U-Net--based autoencoder that generates a binary mask. Both the original image and the explanation are passed through the frozen ResNet, and we tap the post-ReLU activations at five layers along with the final logits. These activations are matched using mean squared error, while the outputs are aligned via KL divergence and cross-entropy. To enforce minimality, we heavily weight the area loss combined with the robustness constraint to generate crisp explanations.

\begin{figure}
    \centering
    \includegraphics[width=0.75\linewidth]{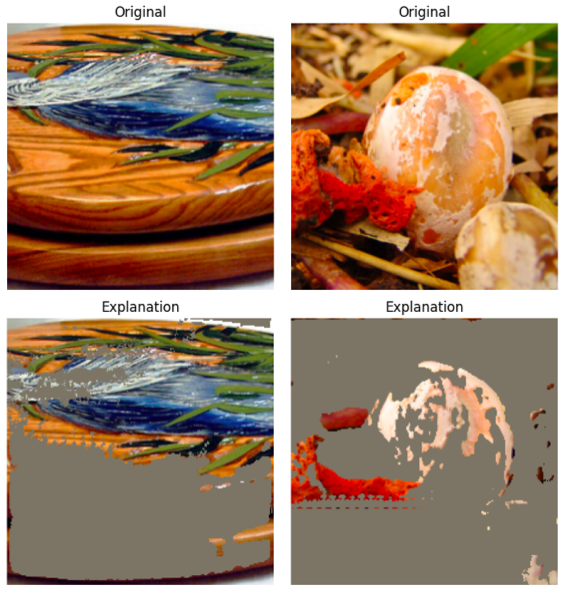}
    \caption{Explanations for misclassified Images. 
    }
    \label{fig:results_cf}
\end{figure}

Figure~\ref{fig:results} illustrates an example for the ImageNet class \emph{EntleBucher}. The first row shows the original image, the binary mask, and the resulting explanation. We observe that the explanation is highly minimal(only about 5\% of active pixels), ignoring background regions of varying colors and textures, and focusing mostly on the object pixels. Interestingly, the top-1 confidence of the explanation is higher than that of the original image, as irrelevant background pixels have been turned off.

As shown in Figure~\ref{fig:results_otters}, in presence strong minimality constraints, the explanation for a single otter reduces to a remarkably small region—roughly 2\% of pixels—focusing primarily on the facial features and fur texture. Despite this extreme sparsity, the classifier’s label is preserved with high confidence. In contrast, when applied to an image with multiple otters, the method produces separate explanations that selectively attend to each animal, demonstrating how the approach can adapt to multi-instance settings and highlight distinct decision-supporting evidence for each occurrence.

\begin{figure*}[t]
    \centering
    \includegraphics[width=\linewidth]{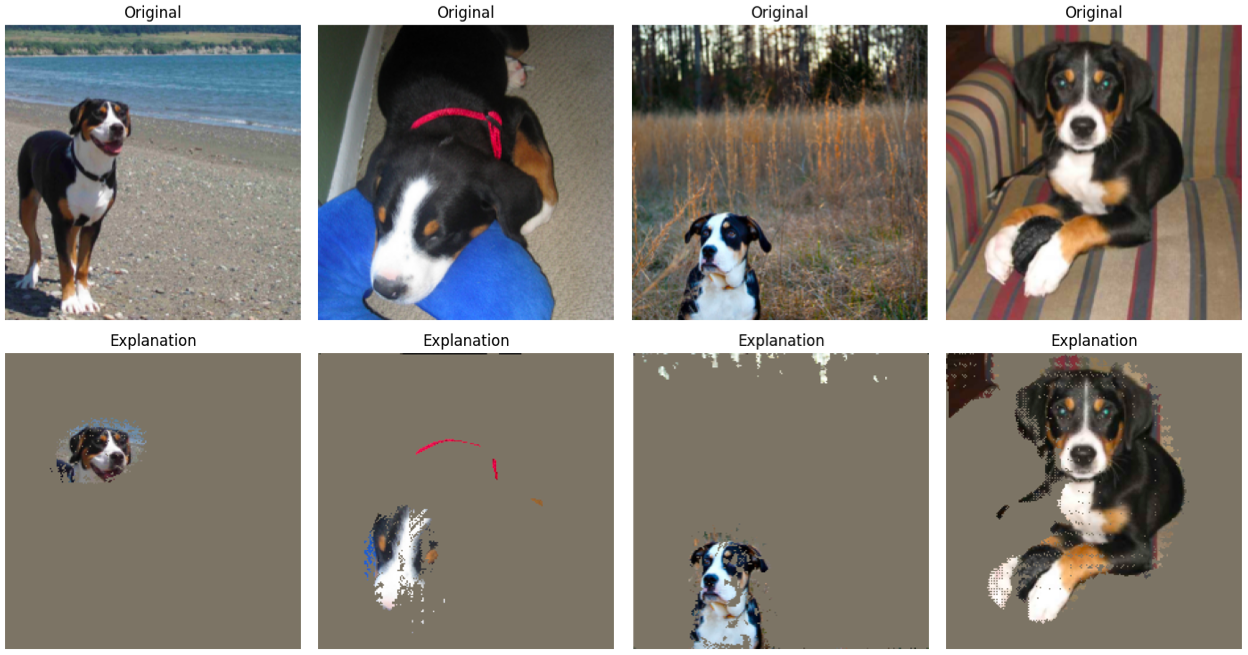}
    \caption{Effect of varying loss weights on generated explanations. 
    (1) With heavily weighted area and total variation losses, the explanation becomes extremely small and localized. 
    (2) Example of shortcut learning: the model highlights not only the dog but also the leash, reflecting dataset biases where dogs frequently appear with leashes. 
    (3) With relaxed constraints, a larger portion of the dog and some background regions are included. 
    (4) Further relaxation of the area loss highlights the entire dog, demonstrating how the approach can be extended toward instance-level segmentation.}
    \label{fig:results_variations}
\end{figure*}

Figure~\ref{fig:results_cf} illustrates how our method sheds light on model errors. 
In the first example, an image of a wooden toilet seat is misclassified as a paint brush. 
The generated explanation reveals that the model focused almost entirely on the decorative print on the seat rather than the seat’s structure, explaining the spurious prediction. 
In another case, an image of a stinkhorn mushroom is misclassified as a goldfish. 
Here, the explanation highlights background regions alongwith the actual fungus, showing that the model mostly ignored the object of interest and instead relied on irrelevant context. 

These examples demonstrate that our explanations not only identify the evidence supporting correct predictions but also expose misleading features responsible for failures, providing actionable insights into model behavior.

Figure~\ref{fig:results_variations} shows how varying the relative weighting of area and smoothness terms affects the explanations. In the first case, heavily weighting the area and total variation losses yields a very compact mask that captures only a small discriminative region. In the second example, the explanation reveals shortcut learning, as the model highlights both the dog and the leash. In the third case, relaxing the minimality constraints results in broader coverage of the dog and partial inclusion of the background. Finally, further relaxation expands the mask to cover the entire object. 

\section{Conclusion}
We presented an activation-matching--based framework for generating minimal and faithful explanations of pretrained image classifiers using an autoencoder that learns to output a mask blocking irrelevant image pixels.

By combining multi-layer activation alignment with output fidelity, mask priors enforcing sparsity and interpretability, and abductive robustness constraints, our approach yields crisp binary masks that preserve both predictions and intermediate activations while discarding irrelevant input regions. These explanations highlight the essential evidence behind a decision, are human-interpretable, and apply equally well to correctly classified and misclassified inputs, thereby helping diagnose model failures. 

Beyond improving per-example explanability, the minimal nature of the explanations can also be used to identify the underlying sparse circuits that realize decisions within deep networks, as part of the future work.

\bibliographystyle{IEEEbib}
\bibliography{strings,refs}

\end{document}